\documentclass[runningheads]{llncs}
\usepackage{graphicx}
\usepackage{amsmath,amssymb} 
\usepackage{color}
\usepackage[width=122mm,left=12mm,paperwidth=146mm,height=193mm,top=12mm,paperheight=217mm]{geometry}
\usepackage[colorlinks=true,bookmarks=false]{hyperref} 
\usepackage{makecell}

\usepackage{tabularx,booktabs}
\newcolumntype{Y}{>{\centering\arraybackslash}X}
\usepackage{floatrow}
\newfloatcommand{capbtabbox}{table}[][\FBwidth]

\begin{document}
\pagestyle{headings}
\mainmatter

\title{Where are the Blobs:\\ Counting by Localization with Point Supervision}
\titlerunning{Where are the Blobs: Counting by Localization with Point Supervision}

\authorrunning{Laradji, Rostamzadeh, Pinheiro, Vazquez, Schmidt}
\author{Issam H. Laradji$^{1,2}$, Negar Rostamzadeh$^{1}$, Pedro O. Pinheiro$^{1}$, David Vazquez$^{1}$, Mark Schmidt$^{2,1}$}
\institute{$^{1}$Element AI, Montreal, Canada\\
           \email{\{negar,pedro,dvazquez\}@elementai.com}\\
           $^{2}$Dept. of Computer Science, University of British Columbia, Vancouver, Canada\\ \email{\{issamou,schmidtm\}@cs.ubc.ca}}

\maketitle

\begin{abstract}
Object counting is an important task in computer vision due to its growing demand in applications such as surveillance, traffic monitoring, and counting everyday objects. State-of-the-art methods use regression-based optimization where they explicitly learn to count the objects of interest. These often perform better than detection-based methods that need to learn the more difficult task of predicting the location, size, and shape of each object. However, we propose a detection-based method that does not need to estimate the size and shape of the objects and that outperforms regression-based methods. Our contributions are three-fold: (1) we propose a novel loss function that encourages the network to output a single blob per object instance using point-level annotations only; (2) we design two methods for splitting large predicted blobs between object instances; and (3) we show that our method achieves new state-of-the-art results on several challenging datasets including the Pascal VOC and the Penguins dataset. Our method even outperforms those that use stronger supervision such as depth features, multi-point annotations, and bounding-box labels. 


\end{abstract}

\section{Introduction}
\label{sec:introduction}
\begin{figure}[t]
\centering
\includegraphics[width=\textwidth]{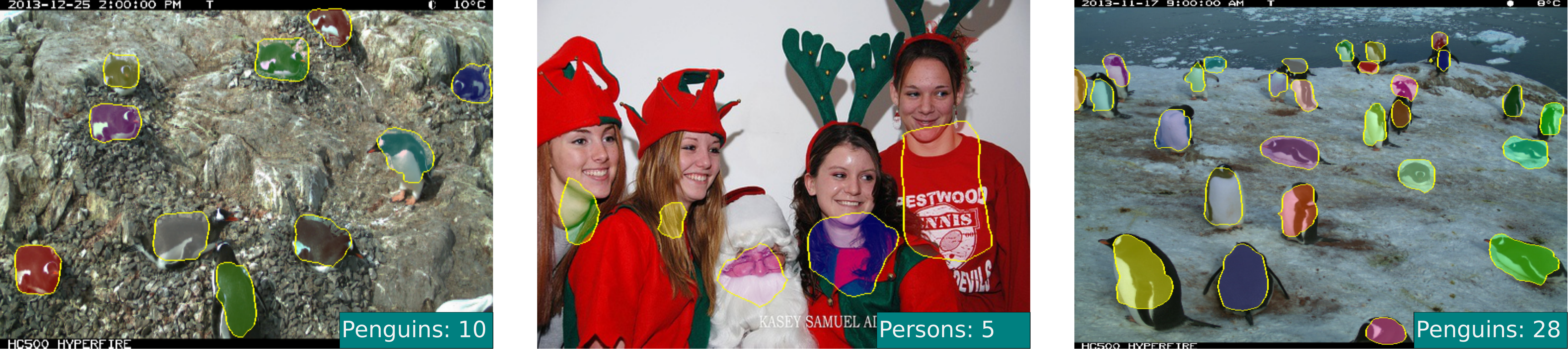}
\caption{Qualitative results on the Penguins~\cite{arteta2016counting} and PASCAL VOC datasets~\cite{everingham2015pascal}. Our method explicitly learns to localize object instances using only point-level annotations. The trained model then outputs blobs where each unique color represents a predicted object of interest. Note that the predicted count is simply the number of predicted blobs.}
\label{fig:res-qualitative}
\end{figure}

Object counting is an important task in computer vision with many applications in surveillance systems~\cite{wang2011automatic,zen2012enhanced}, traffic monitoring~\cite{de2015pklot,TRANCOSdataset_IbPRIA2015}, ecological surveys~\cite{arteta2016counting}, and cell counting~\cite{cohen2017count,lempitsky2010learning}. In traffic monitoring, counting methods can be used to track the number of moving cars, pedestrians, and parked cars. They can also be used to monitor the count of different species such as penguins, which is important for animal conservation. Furthermore, it has been used for counting objects present in everyday scenes in challenging datasets where the objects of interest come from a large number of classes such as the Pascal VOC dataset~\cite{everingham2015pascal}.


Many challenges are associated with object counting. Models need to learn the variability of the objects in terms of shape, size, pose, and appearance. Moreover, objects may appear at different angles and resolutions, and may be partially occluded (see Fig. \ref{fig:res-qualitative}). Also, the background, weather conditions, and illuminations can vary widely across the scenes. Therefore, the model needs to be robust enough to recognize objects in the presence of these variations in order to perform efficient object counting.

Due to these challenges, regression-based models such as ``glance" and object density estimators have consistently defined state-of-the-art results in object counting~\cite{chattopadhyay2016counting,onoro2016towards}. This is because their loss functions are directly optimized for predicting the object count. In contrast, detection-based methods need to optimize for the more difficult task of estimating the location, shape, and size of the object instances. Indeed, perfect detection implies perfect count as the count is simply the number of detected objects. However, models that learn to detect objects often lead to worse results for object counting~\cite{chattopadhyay2016counting}.
For this reason, we look at an easier task than detection by focusing on the  task of simply localizing object instances in the scene. Predicting the exact size and shape of the object instances is not necessary and usually poses a much more difficult optimization problem. Therefore, we propose a novel loss function that encourages the model to output instance regions such that each region contains a single object instance (i.e. a single point-level annotation). Similar to detection, the predicted count is the number of predicted instance regions (see Fig. \ref{fig:res-qualitative}). Our model only requires point supervision which is a weaker supervision than bounding-box, and per-pixel annotations used by most detection-based methods~\cite{ren2015faster,redmon2016you,bai2017deep}. Consequently, we can train our model for most counting datasets as they often have point-level annotations.

This type of annotation is cheap to acquire as it requires lower human effort than bounding box and per-pixel annotations~\cite{bearman2016s}. Point-level annotations provide a rough estimate of the object locations, but not their sizes nor shapes. Our counting method uses the provided point annotations to guide its attention to the object instances in the scenes in order to learn to localize them. As a result, our model has the flexibility to predict different sized regions for different object instances, which makes it suitable for counting objects that vary in size and shape. In contrast, state-of-the-art density-based estimators often assume a fixed object size (defined by the Gaussian kernel) or a constrained environment~\cite{TRANCOSdataset_IbPRIA2015} which makes it difficult to count objects with different sizes and shapes.

Given only point-level annotations, our model uses a novel loss function that (i) enforces it to predict the semantic segmentation labels for each pixel in the image (similar to~\cite{bearman2016s}) and (ii) encourages it to output a segmentation blob for each object instance. During the training phase, the model learns to split the blobs that contain more than one point annotation and to remove the blobs that contain no point-level annotations.

Our experiments show that our method achieves superior object counting results compared to state-of-the-art counting methods including those that use stronger supervision such as per-pixel labels. Our benchmark uses datasets representing different settings for object counting: Mall~\cite{chen2012feature}, UCSD~\cite{chan2008privacy}, and ShanghaiTech B~\cite{zhang2016single} as crowd datasets; MIT Traffic~\cite{wang2009unsupervised}, and Park lot~\cite{de2015pklot} as surveillance datasets; Trancos~\cite{TRANCOSdataset_IbPRIA2015} as a traffic monitoring dataset; and Penguins~\cite{arteta2016counting} as a population monitoring dataset. We also show counting results for the PASCAL VOC~\cite{everingham2015pascal} dataset which consists of objects present in natural, `everyday' images. We also study the effect of using different parts of the proposed loss function against counting and localization performance. 

We summarize our contributions as follows: (1) we propose a novel loss function that encourages the network to output a single blob per object instance using point-level annotations only; (2) we design two methods for splitting large predicted blobs between object instances; and (3) we show that our method achieves new state-of-the-art results on several challenging datasets including the Pascal VOC and the Penguins dataset. 

The rest of the paper is organized as follows: Section~\ref{sec:related-works} presents related works on object counting; Section~\ref{sec:loss} describes the proposed approach; and Section~\ref{sec:exp} describes our experiments and results. Finally, we present the conclusion in Section~\ref{sec:conclusion}.

\section{Related Work}
\label{sec:related-works}

Object counting has received significant attention over the past years~\cite{rabaud2006counting,chattopadhyay2016counting,lempitsky2010learning}. It can be roughly divided into three categories~\cite{loy2013crowd}: (1) counting by clustering, (2) counting by regression, and (3) counting by detection.

Early work in object counting use {\it clustering-based methods}. They are unsupervised approaches where objects are clustered based on features such as appearance and motion cues~\cite{rabaud2006counting,tu2008unified}. Rabaud and Belongie~\cite{rabaud2006counting} proposed to use feature points which are detected by motion and appearance cues and are tracked through time using KLT~\cite{Shi1993}. The objects are then clustered based on similar features. Sebastian \emph{et al.}~\cite{tu2008unified} used an expectation-maximization method that cluster individuals in crowds based on head and shoulder features. These methods use basic features and often perform poorly for counting compared to deep learning approaches. Another drawback is that these methods only work for video sequences, rather than still images.

\textit{Counting by regression} methods have defined state-of-the-art results in many benchmarks. They were shown to be faster and more accurate than other groups such as counting by detection. These methods include glance and density-based methods that explicitly learn how to count rather than optimize for a localization-based objective. Lempitsky \emph{et al.}~\cite{lempitsky2010learning} proposed the first method that used object density to count people. They transform the point-level annotation matrix into a density map using a Gaussian kernel. Then, they train their model using a least-squares objective to predict the density map. One major challenge is determining the optimal size of the Gaussian kernel which highly depends on the object sizes. As a result, Zhang \emph{et al.}~\cite{zhang2016single} proposed a deep learning method that adjusted the kernel size using a perspective map. This assumes fixed camera images such as those used in surveillance applications. Onoro-Rubio \emph{et al.}~\cite{onoro2016towards}  extended this method by proposing a perspective-free multi-scale deep learning approach. However, this method cannot be used for counting everyday objects as their sizes vary widely across the scenes as it is highly sensitive to the kernel size.

A straight-forward method for counting by regression is `glance'~\cite{chattopadhyay2016counting}, which explicitly learns to count using image-level labels only. Glance methods are efficient if the object count is small~\cite{chattopadhyay2016counting}. Consequently, the authors proposed a grid-based counting method, denoted as ``subitizing", in order to  count a large number of objects in the image.  This method uses glance to count objects at different non-overlapping regions of the image, independently. While glance is easy to train and only requires image-level annotation, the ``subitizing" method requires a more complicated training procedure that needs full per-pixel annotation ground-truth.

\textit{Counting by detection} methods first detect the objects of interest and then simply count the number of instances. Successful object detection methods rely on bounding boxes~\cite{ren2015faster,redmon2016you,liu2016ssd} and per-pixel labels~\cite{long2015fully,jegou2017one,zhao2017pyramid} ground-truth.  Perfect object detection implies perfect count. However, Chattopadhyay \emph{et al.}~\cite{chattopadhyay2016counting} showed that Fast RCNN~\cite{girshick2015fastrcnn}, a state-of-the-art object detection method, performs worse than glance and subitizing-based methods. This is because the detection task is challenging in that the model needs to learn the location, size, and shape of object instances that are possibly heavily occluded. While several works~\cite{chattopadhyay2016counting,onoro2016towards,lempitsky2010learning} suggest that counting by detection is infeasible for surveillance scenes where objects are often occluded, we show that learning a notion of localization can help the model improve counting.

Similar to our method is the line of work proposed by Arteta \emph{et al.}~\cite{Arteta2012,Arteta2013,Arteta2016c}. They proposed a method that detects overlapping instances based on optimizing a tree-structured discrete graphical model. While their method showed promising detection results using point-level annotations only, it performed worse for counting than regression-based methods such as~\cite{lempitsky2010learning}. 

Our method is also similar to segmentation methods such as U-net~\cite{Ronneberger2015} which learns to segment objects using a fully-convolutional neural network. Unlike our method, U-net requires the full per-pixel instance segmentation labels, whereas we use point-level annotations only.

\section{Localization-based Counting FCN}
\label{sec:loss}

Our model is based on the fully convolutional neural network (FCN) proposed by Long \emph{et al.}~\cite{long2015fully}. We extend their semantic segmentation loss to perform object counting and localization with point supervision. We denote the novel loss function as \emph{localization-based counting loss} (LC) and, we refer to the proposed model as LC-FCN. Next, we describe the proposed loss function, the architecture of our model, and the prediction procedure.


\subsection{The Proposed Loss Function}
LC-FCN uses a novel loss function that consists of four distinct terms. The first two terms, the image-level and the point-level loss, enforces the model to predict the semantic segmentation labels for each pixel in the image. This is based on the weakly supervised semantic segmentation algorithm proposed by Bearman \emph{et al.}~\cite{bearman2016s}. These two terms alone are not suitable for object counting as the predicted blobs often group many object instances together (see the ablation studies in Section \ref{sec:exp}). The last two terms encourage the model to output a unique blob for each object instance and remove blobs that have no object instances. Note that LC-FCN only requires point-level annotations that indicate the locations of the objects rather than their sizes, and shapes.

Let $T$ represent the point annotation ground-truth matrix which has label $c$ at the location of each object (where $c$ is the object class) and zero elsewhere. Our model uses a  \emph{softmax} function to output a matrix $S$ where each entry $S_{ic}$ is the probability that pixel $i$ belongs to category $c$. The proposed loss function can be written as:
\begin{equation}
\mathcal{L}(S, T) = \underbrace{\mathcal{L}_I(S,T)}_{\text{Image-level loss}} + \underbrace{\mathcal{L}_P(S,T)}_{\text{Point-level loss}} + \underbrace{\mathcal{L}_S(S,T)}_{\text{Split-level loss}} + \underbrace{\mathcal{L}_F(S,T)}_{\text{False positive loss}}\;,
\label{eq:loss}
\end{equation}
which we describe in detail next.

\subsubsection{Image-level loss.} Let $C_e$ be the set of classes present in the image. For each class $c \in C_e$, $\mathcal{L}_I$ increases the probability that the model labels at least one pixel as class $c$. Also, let $C_{\neg{e}}$ be the set of classes not present in the image. For each class $c \in C_{\neg{e}}$, the loss decreases the probability that the model labels any pixel as class $c$. $C_e$ and $C_{\neg{e}}$ can be obtained from the provided ground-truth point-level annotations. More formally, the image level loss is computed as follows:
\begin{equation}
\mathcal{L}_I(S, T) = -\frac{1}{|C_e|}\sum_{c\in C_e}\log(S_{t_cc}) -\frac{1}{|C_{\neg{e}}|}\sum_{c\in C_{\neg{e}}}\log(1 - S_{t_cc}) \;,
\label{eq:imagelevel}
\end{equation}
where $t_c = \text{argmax}_{i \in \mathcal{I}} S_{ic}$.
For each category present in the image, at least one pixel should be labeled as that class. For classes that do not exist in the image, none of the pixels should belong to that class. Note that we assume that each image has at least one background pixel; therefore, the background class belongs to $C_e$.

\subsubsection{Point-level loss.}This  term encourages the model to correctly label the small set of supervised pixels $\mathcal{I}_s$ contained in the ground-truth. $\mathcal{I}_s$ represents the locations of the object instances. This is formally defined as,
\begin{equation} \label{eq:pointlevel}
\mathcal{L}_P(S, T) = -\sum_{i\in \mathcal{I}_s}\log(S_{iT_i})\;,
\end{equation}
where $T_i$ represents the true label of pixel $i$. Note that this loss ignores all the pixels that are not annotated.

\subsubsection{Split-level loss.} $\mathcal{L}_S$ discourages the model from predicting blobs that have two or more point-annotations. Therefore, if a blob has $n$ point annotations, this loss enforces it to be split into $n$ blobs, each corresponding to a unique object. These splits are made by first finding boundaries between object pairs. The model then learns to predict these boundaries as the background class. The model outputs a binary matrix $\mathcal{F}$ where pixel $i$ is foreground if $\text{argmax}_k S_{ik} > 0$, and background, otherwise.

\begin{figure}[!t]
\centering
\includegraphics[width=1.0\textwidth]{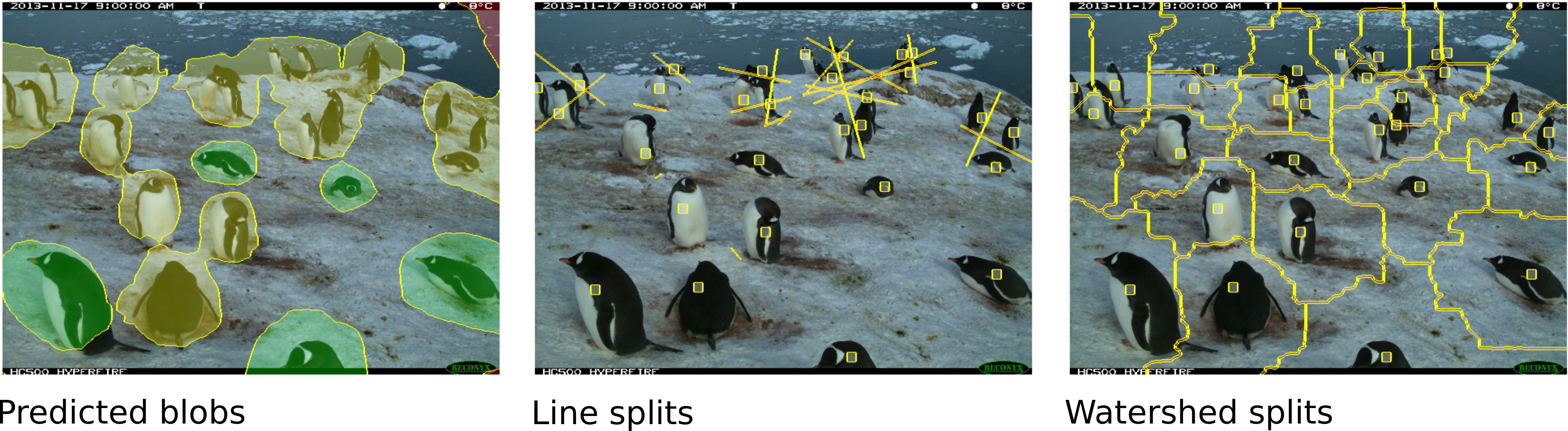}
\caption{{\bf Split methods.} Comparison between the line split, and the watershed split. The loss function identifies the boundary splits (shown as yellow lines). Yellow blobs represent those with more than one object instance, and red blobs represent those that have no object instance. Green blobs are true positives. The squares represent the ground-truth point annotations.}
\label{fig:splits}
\end{figure}

We apply the connected components algorithm proposed by~\cite{wu2005optimizing} to find the blobs $B$ in the foreground mask $\mathcal{F}$. We only consider the blobs with two or more ground truth point annotations $\bar{B}$. We propose two methods for splitting blobs (see Fig. \ref{fig:splits}),

\begin{enumerate}
\item {\it Line split method}. For each blob $b$ in $\bar{B}$ we pair each point with its closest point resulting in a set of pairs $b_P$. For each pair $(p_i, p_j) \in b_P$ we use a scoring function to determine the best segment $E$ that is perpendicular to the line between $p_i$ and $p_j$. The segment lines are within the predicted blob and they intersect the blob boundaries. The scoring function $z(\cdot)$ for segment E is computed as,
\begin{equation}
z(E) = \frac{1}{|E|}\sum_{i \in E}S_{i0}\;,
\end{equation}
which is the mean of the background probabilities belonging to segment $E$ (where 0 is the background class). The best edge $E_{best}$ is defined as the set of pixels representing the edge with the highest probability of being background among all the perpendicular lines. This determines the `most likely' edge of separation between the two objects. Then we set $T_b$ as the set of pixels representing the best edges generated by the line split method.


\item {\it Watershed split method}. This consists of global and local segmentation procedures. For the global segmentation, we apply the watershed segmentation algorithm \cite{beucher1992morphological} globally on the input image where we set the ground-truth point-annotations as the seeds.  The segmentation is applied on the distance transform of the foreground probabilities, which results in $k$ segments where $k$ is the number of point-annotations in the image.

For the local segmentation procedure, we apply the watershed segmentation only within each blob $b$ in $\bar{B}$ where we use  the point-annotation ground-truth inside them as seeds. This adds more importance to splitting big blobs when computing the loss function. Finally, we define $T_b$ as the set of pixels representing the boundaries determined by the local and global segmentation.
\end{enumerate}

Fig. \ref{fig:splits} shows the split boundaries using the line split and the watershed split methods  (as yellow lines). Given $T_b$, we compute the split loss as follows,

\begin{equation}
\begin{split}
\mathcal{L}_S(S, T) &=  -  \sum_{i \in T_b} \alpha_i \log(S_{i0}),
\end{split}
\label{eq:edgelevel}
\end{equation}
where $S_{i0}$ is the probability that pixel $i$ belongs to the background class and $\alpha_i$ is the number of point-annotations in the blob in which pixel $i$ lies. This encourages the model to focus on splitting blobs that have the most point-level annotations. The intuition behind this method is that learning to predict the boundaries between the object instances allows the model to distinguish between them. As a result, the penalty term encourages the model to output a single blob per object instance. 

We emphasize that it is not necessary to get the right edges in order to accurately count. It is only necessary to make sure we have a positive region on each object and a negative region between objects. Other heuristics are possible to construct a negative region which could still be used in our framework. For example, fast label propagation methods proposed in \cite{nutini2017let,nutini2016convergence} can be used to determine the boundaries between the objects in the image. Note that these 4 loss functions are only used during training. The framework does not split or remove false positive blobs at test time. The predictions are based purely on the blobs obtained from the probability matrix $S$.



\subsubsection{False Positive loss.} $\mathcal{L}_F$ discourages the model from predicting a blob with no point annotations, in order to reduce the number of false positive predictions. The loss function is defined as

\begin{equation}
\mathcal{L}_F(S, T) = - \sum_{i \in B_{fp}} \log(S_{i0}),
\label{eq:falsepositive}
\end{equation}
where $B_{fp}$ is the set of pixels constituting the blobs predicted for each class (except the background class) that contain no ground-truth point annotations (note that $S_{i0}$ is the probability that pixel $i$ belongs to the background class). All the predictions within $B_{fp}$ are considered false positives (see the red blobs in Fig. \ref{fig:ablation}). Therefore, optimizing this loss term results in less false positive predictions as shown in the qualitative results in Fig. \ref{fig:ablation}. The experiments show that this loss term is extremely important for accurate object counting.

\subsection{LC-FCN Architecture and Inference}
\begin{figure}[!t]
\centering
\includegraphics[width=1.0\textwidth]{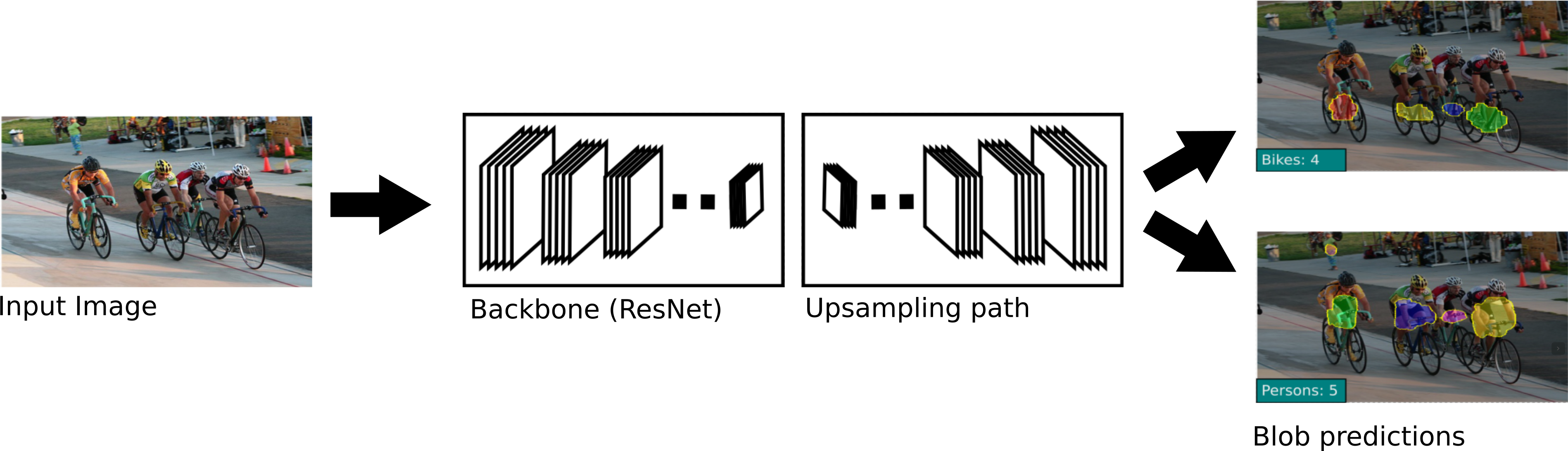}
\caption{Given an input image, our model first extracts features using a backbone architecture such as ResNet. The extracted features are then upsampled through the upsampling path to obtain blobs for the objects. In this example, the model predicts the blobs for persons and bikes for an image in the PASCAL VOC 2007 dataset.}
\label{fig:model}
\end{figure}

LC-FCN can be any FCN architecture such as FCN8 architecture~\cite{long2015fully}, Deeplab~\cite{chen2016deeplab}, Tiramisu~\cite{jegou2017one}, and PSPnet~\cite{zhao2017pyramid}. LC-FCN consists of a backbone that extracts the image features. The backbone is an Imagenet pretrained network such as VGG16 or ResNet-50~\cite{simonyan2014very,imagenet_cvpr09}. The image features are then upscaled using an upsampling path to output a score for each pixel $i$ indicating the probability that it belongs to class $c$ (see Fig. \ref{fig:model}). 

We predict the number of objects for class $c$ through the following three steps: (i) the upsampling path outputs a matrix $Z$ where each entry $Z_{ic}$ is the probability that pixel $i$ belongs to class $c$; then (ii) we generate a binary mask $F$, where pixel $F_{ic}=1$ if $\arg \max_k Z_{ik}=c$, and 0 otherwise; lastly (iii) we apply the connected components algorithm~\cite{wu2005optimizing} on $F$ to get the blobs for each class $c$. The count is the number of predicted blobs (see Fig. \ref{fig:model}).


\begin{table}[!ht]
\centering
\caption{{\bf Penguins datasets.} Evaluation of our method against previous state-of-the-art methods. The evaluation is made across the four setups explained in the dataset description.}
\def\tabularxcolumn#1{m{#1}}
\begin{tabularx}{\textwidth}{l*{3}{Y|}*{1}{Y}}
  \multicolumn{1}{c}{} & \multicolumn{2}{c|}{\bf Separated}  & \multicolumn{2}{c}{\bf Mixed}\\ \hline
  {\bf Method }  & Max &Median & Max & Median\\ \hline\hline
  Density-only~\cite{arteta2016counting} & 8.11 & 5.01&9.81&7.09\\  \hline
  With seg. and depth~\cite{arteta2016counting} & 6.38  & 3.99&5.74&3.42\\ \hline
  With seg and no depth~\cite{arteta2016counting} & 5.77  & 3.41&5.35&3.26\\ \hline\hline
  Glance & 6.08  & 5.49 &1.84&  2.14\\ \hline
  LC-FCN8& {\bf 3.74}  & {\bf 3.28}& 1.62& 1.80\\ \hline
  LC-ResFCN& 3.96 & 3.43 &  {\bf 1.50} &  {\bf 1.69}\\ \hline
\end{tabularx}
\label{table:penguins}
\end{table}
\section{Experiments}
\label{sec:exp}
In this section we describe the evaluation metrics, the training procedure, and present the experimental results and discussion.

\subsection{Setup}
\subsubsection{Evaluation Metric.}
For datasets with single-class objects, we report the mean absolute error (MAE) which measures the deviation of the predicted count $p_i$ from the true count $c_i$, computed as $\frac{1}{N}\sum_i|p_i - c_i|$. MAE is a commonly used metric for evaluating object counting methods~\cite{charles2015nonnegative,sindagi2017survey}. For datasets with multi-class objects, we report the mean root mean square error (mRMSE) as used in~\cite{chattopadhyay2016counting}
for the PASCAL VOC 2007 dataset. We measure the localization performance using the average mean absolute error (GAME) as in~\cite{TRANCOSdataset_IbPRIA2015}. Since our model predicts blobs instead of a density map, GAME might not be an accurate localization measure. Therefore, in section \ref{sec:ablation} we use the F-Score metric to assess the localization performance of the predicted blobs against the point-level annotation ground-truth.

\subsubsection{Training Procedure.}
We use the Adam~\cite{kingma2014adam} optimizer with a learning rate of $10^{-5}$ and weight decay of $5\times10^{-5}$. We use the provided validation set for early stopping only. During training, the model uses a batch size of 1 which can be an image of any size. We double our training set by applying the horizontal flip augmentation method on each image. Finally, we report the prediction results on the test set. We compare between three architectures: FCN8~\cite{long2015fully}; ResFCN which is FCN8 that uses ResNet-50 as the backbone instead of VGG16; and PSPNet~\cite{zhao2017pyramid} with ResNet-101 as the backbone. We use the watershed split procedure in all our experiments.

\subsection{Results and Discussion}\label{sec:results}

\subsubsection{Penguins Dataset~\cite{arteta2016counting}.}
The Penguins dataset comprises images of penguin colonies located in Antarctica. We use the two dataset splits as in~\cite{arteta2016counting}. In the `separated' dataset split, the images in the training set come from different cameras than those in the test set. In the `mixed' dataset split, the images in the training set come from the same cameras as those in the test set. In Table \ref{table:penguins}, the MAE is computed with respect to the Max and Median count (as there are multiple annotators). Our methods significantly outperform theirs in all of the four settings, although their methods use depth features and the multiple annotations provided for each penguin. This suggests that LC-FCN can learn to distinguish between individual penguins despite the heavy occlusions and crowding.

\subsubsection{Trancos Dataset~\cite{onoro2016towards}.}\label{sec:trancos}
The Trancos dataset comprises images taken from traffic surveillance cameras located along different roads. The task is to count the vehicles present in the regions of interest of the traffic scenes. Each vehicle is labeled with a single point annotation that represents its location in the image.
We observe in Table \ref{table:trancos} that our method achieves new state-of-the-art results for counting and localization. Note that $GAME(L)$ subdivides the image using a grid of $4^L$ non-overlapping regions, and the error is computed as the sum of the mean absolute errors in each of these subregions. For our method, the predicted count of a region is the number of predicted blob centers in that region. This provides a rough assessment of the localization performance. Compared to the methods in Table \ref{table:trancos}, LC-FCN does not require a perspective map nor a multi-scale approach to learn objects of different sizes. These results suggest that LC-FCN can accurately localize and count extremely overlapping vehicles.

\begin{table}[t]
\centering
\caption{{\bf Trancos dataset.} Evaluation of our method against previous state-of-the-art methods, comparing the mean absolute error (MAE) and the grid average mean absolute error (GAME) as described in~\cite{TRANCOSdataset_IbPRIA2015}.}
\def\tabularxcolumn#1{m{#1}}
\begin{tabularx}{\textwidth}{l *{3}{Y|}*{1}{Y} }
  {\bf Method} & {\bf MAE} & {\bf GAME(1)} & {\bf GAME(2)} & {\bf GAME(3)} \\\hline\hline
  Lemptisky+SIFT~\cite{TRANCOSdataset_IbPRIA2015} & 13.76&16.72&20.72&24.36 \\\hline
  Hydra CCNN~\cite{onoro2016towards}  & 10.99 &13.75&16.69&19.32\\\hline
   FCN-MT~\cite{zhang2017understanding}  & 5.31 &-&-&-\\\hline
 FCN-HA~\cite{zhang2017fcn}  & 4.21 &-&-&-\\\hline
  CSRNet~\cite{Yuhong2018}  & 3.56 &5.49&8.57&15.04\\\hline\hline
  Glance & 7.0  & -  & - &  -\\\hline
  LC-FCN8  &  4.53  &7.00 & 10.66 & 16.05 \\\hline
    LC-ResFCN & {\bf 3.32}  &5.2 &7.92 &12.57 \\\hline
    LC-PSPNET &  3.57  &{\bf 4.98} &{\bf 7.42} &{\bf 11.67} \\\hline
\end{tabularx}
\label{table:trancos}
\end{table}

\begin{table}
\centering
\caption{{\bf PASCAL VOC.} We compare against the methods proposed in ~\cite{chattopadhyay2016counting}. Our model evaluates on the full test set, whereas the other methods take the mean of ten random samples of the test set evaluation.}

\begin{tabular}{l|c|c|c|c}
  \bf Method & \bf mRMSE & \bf mRMSE-nz & \bf m-relRMSE & \bf {\small m-relRMSE-nz}\\ 
 \hline\hline
  Glance-noft-2L~\cite{chattopadhyay2016counting}& $0.50 $& $1.83 $ & $0.27 $ & $0.73$\\ 
   \hline
   Aso-sub-ft-$3\times 3$~\cite{chattopadhyay2016counting} & $0.42$ & $1.65 $ & $0.21 $ & $0.68 $\\ \hline 
  Faster-RCNN~\cite{chattopadhyay2016counting}& $0.50$ & $1.92$ & $0.26$&$0.85$\\
 \hline
 \hline
LC-ResFCN& {\bf 0.31} & {\bf 1.20}&{\bf 0.17}&   {\bf 0.61}\\ 
\hline
LC-PSPNet& 0.35 & 1.32 & 0.20&  0.70\\ 
\hline
\end{tabular}
\label{tab:pascal}
\end{table}

\subsubsection{Parking Lot~\cite{de2015pklot}.}
The dataset comprises surveillance images taken at a parking lot in Curitiba, Brazil. We used the PUCPR subset of the dataset where the first 50\% of the images was set as the training set and the last 50\% as the test set. The last 20\% of the training set was set as the validation set for early stopping. The ground truth consists of a bounding box for each parked car since this dataset is primarily used for the detection task. Therefore, we convert them into point-level annotations by taking the center of each bounding box. Table \ref{table:ablation} shows that LC-FCN significantly outperforms Glance in MAE. LC-FCN8 achieves only 0.21 average miscount per image although many images contain more than 20 parked cars. This suggests that explicitly learning to localize parked cars can perform better in counting than methods that explicitly learn to count from image-level labels (see Fig. \ref{fig:ablation} for qualitative results). Note that this is the first counting method being applied on this dataset.

\vspace{-.5cm}
\subsubsection{MIT Traffic~\cite{wang2011automatic}.} This dataset consists of surveillance videos taken from a single fixed camera. It has 20 videos, which are split into a training set (Videos 1-8), a validation set (Videos 0-10), and a test set (Videos 11-20). Each video frame is provided with a bounding box indicating each pedestrian. We convert them into point-level annotations by taking the center of each bounding box. Table \ref{table:ablation} shows that our method significantly outperforms Glance, suggesting that learning a localization-based objective allows the model to ignore the background regions that do not contribute to the object count. As a result, LC-FCN is less likely to overfit on irrelevant features from the background. To the best of our knowledge, this is the first counting method being applied on this dataset.

\begin{figure}[t]
\begin{floatrow}
\capbtabbox{
  \resizebox{.45\textwidth}{!}{
    \begin{tabular}{ l | c | c | c   }
      Methods & UCSD &  Mall & ShanghaiTech B  \\ \hline
      FCN-rLSTM~\cite{zhang2017fcn}& 1.54 & - & -\\  \hline
      MoCNN~\cite{kumagai2017mixture}& - & 2.75 &-\\  \hline
      CNN-boosting~\cite{walach2016learning} & 1.10 & 2.01 & -\\\hline
      M-CNN~\cite{zhang2016single} & 1.07 & - & 26.4\\ \hline
      CP-CNN ~\cite{sindagi2017generating}& - &  - & 20.1\\ \hline
       CSRNet ~\cite{Yuhong2018} & 1.16 &  - & {\bf 10.6}\\\hline\hline
      LC-FCN8& 1.51 &  2.42&   13.14\\ \hline
      LC-ResFCN& {\bf 0.99} & 2.12 &  25.89\\ \hline
      LC-PSPNet& 1.01 & {\bf 2.00} &  21.61\\ \hline
    \end{tabular}
 }
}
{\caption{Crowd datasets MAE results.} \label{tab:crowd}}
\ffigbox{
  \includegraphics[height=.3\textwidth]{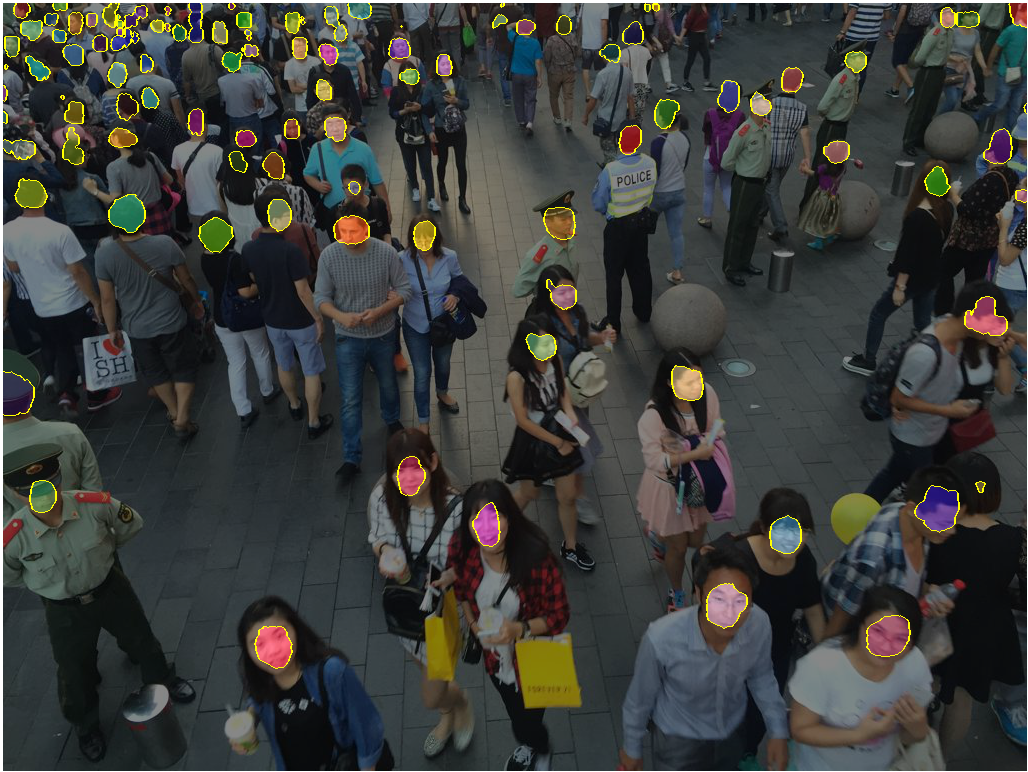}}
{\caption{Predicted blobs on a ShanghaiTech B test image.} \label{fig:qual_shanghai}}
\end{floatrow}
\end{figure}\textbf{}

\vspace{-.6cm}
\subsubsection{Pascal VOC 2007~\cite{everingham2015pascal}.}  We use the standard training, validation, and test split as specified in~\cite{everingham2015pascal}. We use the point-level annotation ground-truth provided by Bearman \emph{et al.}~\cite{bearman2016s} to train our LC-FCN methods. We evaluated against the count of the non-difficult instances of the Pascal VOC 2007 test set.


Table \ref{tab:pascal} compares the performance of LC-FCN with different methods proposed by~\cite{chattopadhyay2016counting}. We point the reader to~\cite{chattopadhyay2016counting} for a description of the evaluation metrics used in the table. We show that LC-FCN achieves new state-of-the-art results with respect to mRMSE. We see that LC-FCN outperforms methods that explicitly learn to count although learning to localize objects of this dataset is a very challenging task. Further, LC-FCN uses weaker supervision than Aso-sub and Seq-sub as they require the full per-pixel labels to estimate the object count for different image regions.

\subsubsection{Crowd Counting Datasets.}
Table \ref{tab:crowd} reports the MAE score of our method on 3 crowd datasets using the setup described in the survey paper~\cite{sindagi2017survey}.  For this experiment, we show our results using ResFCN as the backbone with the Watershed split method. We see that our method achieves competitive performance for crowd counting. 
Fig. \ref{fig:qual_shanghai} shows the predicted blobs of our model on a test image of the ShanghaiTech B dataset. We see that our model predicts a blob on the face of each individual. This is expected since the ground-truth point-level annotations are marked on each person's face.

\begin{table}[!t]
\caption{{\bf Quantitative results.}  Comparison of different parts of the proposed loss function for counting and localization performance.}
\begin{center}
\def\tabularxcolumn#1{m{#1}}
\begin{tabularx}{\textwidth}{l *{7}{Y|}*{1}{Y} }
  \multicolumn{1}{c}{} & \multicolumn{2}{c|}{\bf MIT Traffic} & \multicolumn{2}{c|}{\bf PKLot} & \multicolumn{2}{c|}{\bf Trancos} &\multicolumn{2}{c}{\bf \makecell[c]{Penguins \\Separated} } \\\hline
  \bf Method  & \bf MAE & \bf FS & \bf MAE & \bf FS & \bf MAE & \bf FS & \bf MAE & \bf FS\\\hline\hline
  Glance & 1.57 &-&1.92&-&7.01&-&6.09&-\\\hline
  \makecell[l]{$\mathcal{L}_I + \mathcal{L}_P$} &3.11 &0.38&39.62&0.04&38.56&0.05&9.81&0.08\\\hline
  \makecell[l]{$\mathcal{L}_I + \mathcal{L}_P + \mathcal{L}_S$ }& 1.62&0.76&9.06&0.83&6.76&0.56&4.92&0.53\\\hline
   \makecell[l]{$\mathcal{L}_I + \mathcal{L}_P + \mathcal{L}_F$} &1.84 &0.69&39.60&0.04&38.26&0.05&7.28&0.04 \\\hline
   LC-ResFCN &1.26 &{\bf 0.81}& 10.16&0.84&{\bf 3.32}&0.68& 3.96& 0.63\\\hline
      LC-FCN8 & {\bf 0.91} & 0.69&{\bf 0.21}&{\bf0.99}&4.53&{\bf0.54}&{\bf 3.74}&{\bf 0.61}\\\hline

\end{tabularx}
\end{center}
\label{table:ablation}
\end{table}

\begin{figure}[ht]
\centering
\includegraphics[width=\textwidth]{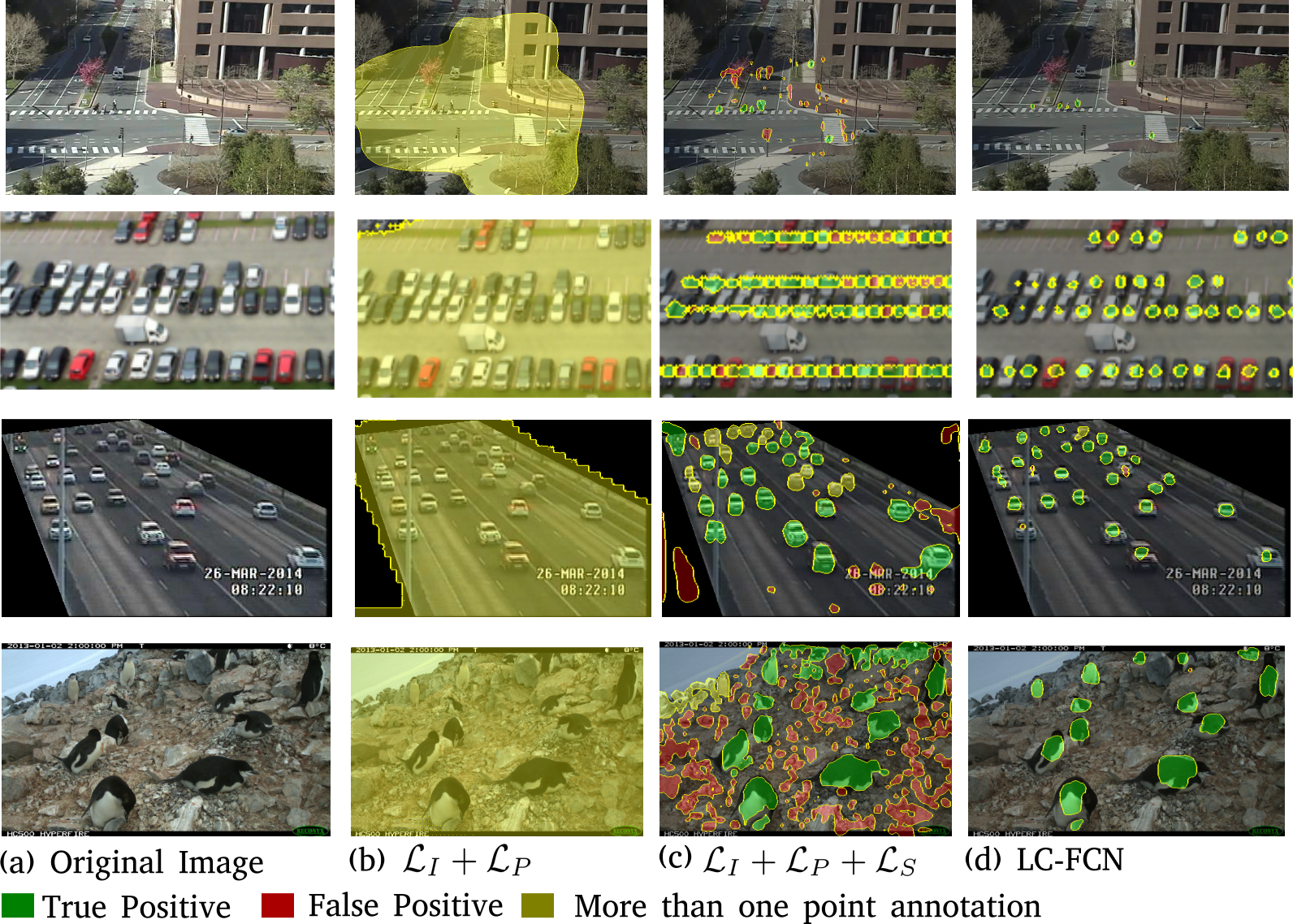}
\caption{Qualitative results of LC-FCN trained with different terms of the proposed loss function. (a) Test images obtained from MIT Traffic,  Parking Lot, Trancos, and Penguins. (b) Prediction results using only image-level and point-level loss terms. (c) Prediction results using image-level, point-level, and split-level loss terms. (d) Prediction results trained with the full proposed loss function. The green blobs and red blobs indicate true positive and false positive predictions, respectively. Yellow blobs represent those that contain more than one object instance.}
\label{fig:ablation}
\end{figure}

\subsection{Ablation Studies}\label{sec:ablation}

\subsubsection{Localization Benchmark.} Since robust localization is useful in many computer vision applications, we use the F-Score measure to directly assess the localization performance of our model. F-Score is a standard measure for detection as it considers both precision and recall, $\text{F-Score} =\frac{2\text{TP}}{2\text{TP} + \text{FP} + \text{FN}}$, where the number of true positives (TP) is the number of blobs that contain at least one point annotation; the number of false positives (FP) is the number of blobs that contain no point annotation; and the number of false negatives (FN) is the number of point annotations minus the number of true positives. Table \ref{table:ablation} shows the localization results of our method on several datasets.



\begin{figure}
\caption{{\bf Split Heuristics Analysis.} Comparison between the watershed split method and the line split method against the validation MAE score.}
\label{fig:crowd2}
\hspace{-0.35in}
\begin{floatrow}
\capbtabbox{%
  \begin{tabular}{ l | c | c  }
 {\bf Split method} & {\bf Trancos }& {\bf Penguins} \\ \hline
LC-ResFCN (L) &4.77 &1.89\\ 
\hline
LC-ResFCN (W) & 3.34&0.95\\ 
\hline
\end{tabular}
}
{\label{tab:res-pascal}}
\hspace{-2.8in}
\ffigbox{
  \includegraphics[width=.26\textwidth]{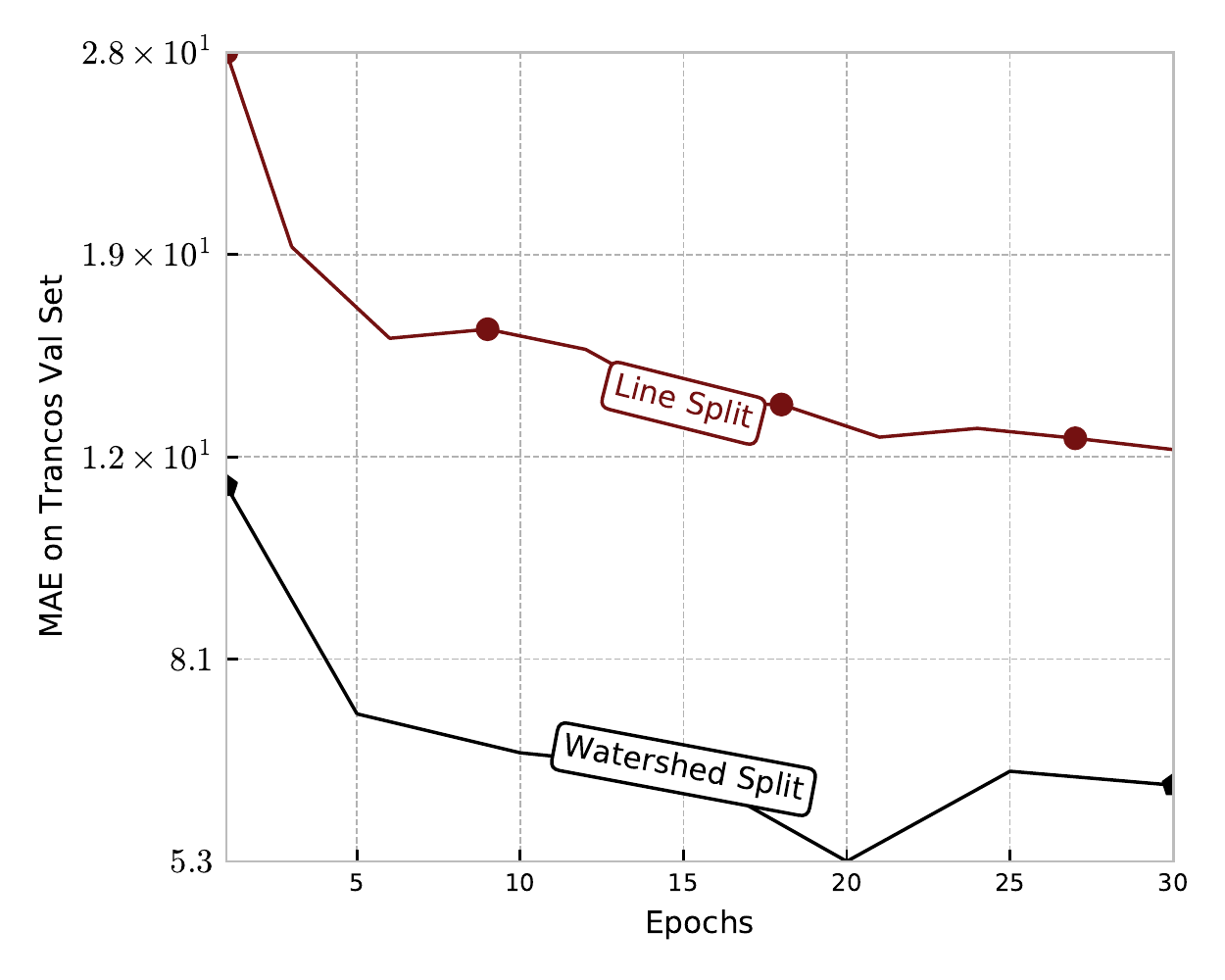}~
  \includegraphics[width=.26\textwidth]{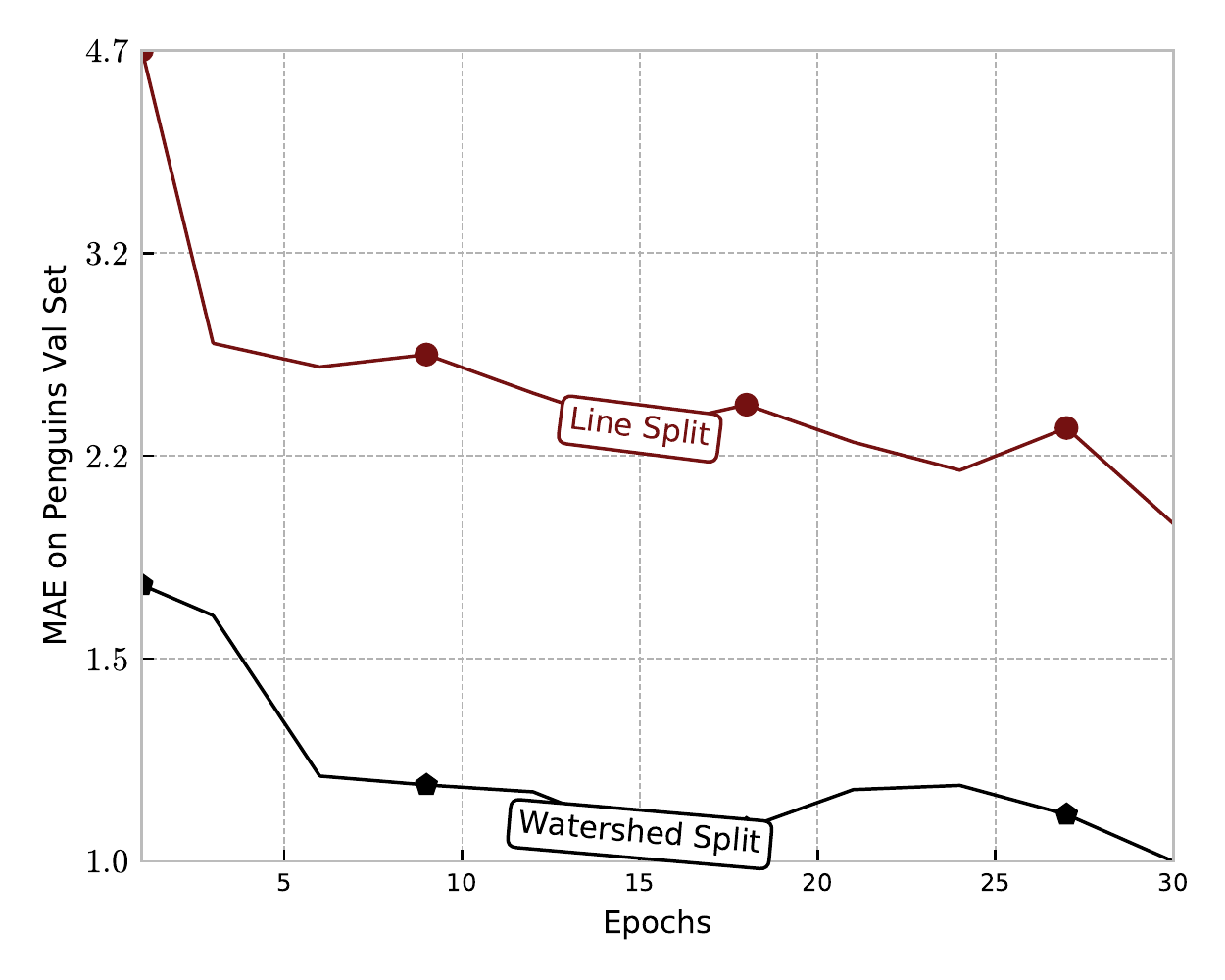}\vspace{-1mm}
}{\label{fig:speed}}
\end{floatrow}
\end{figure}
\subsubsection{Loss Function Analysis.}
We assess the effect of each term of the loss function on counting and localization results. We start by looking at the results of a model trained with the image-level loss $\mathcal{L}_I$ and the point-level loss $\mathcal{L}_P$ only. These two terms were used for semantic segmentation using point annotations~\cite{bearman2016s}. We observe in Fig.~\ref{fig:ablation}(b) that a model using these two terms results in a single blob that groups many object instances together. Consequently, this performs poorly in terms of the mean absolute error and the F-Score (see Table \ref{table:ablation}).  As a result, we introduced the split-level loss function $\mathcal{L}_S$ that encourages the model to predict blobs that do not contain more than one point-annotation. We see in Fig.~\ref{fig:ablation}(c) that a model using this additional loss term predicts several blobs as object instances rather than one large single blob. However,  since $\mathcal{L}_I + \mathcal{L}_P  + \mathcal{L}_S$ does not penalize the model from predicting blobs with no point annotations, it can often lead to many false positives. Therefore, we introduce the false positive loss $\mathcal{L}_F$ that discourages the model from predicting blobs with no point annotations. By adding this loss term to the optimization, LC-FCN achieves significant improvement as seen in the qualitative and quantitative results (see Fig.~\ref{fig:ablation}(d) and Table~\ref{table:ablation}). Further, including only the split-level loss leads to predicting a huge number of small blobs, leading to many false positives which makes performance worse. Combining it with the false-positive loss avoids this issue which leads to a net improvement in performance. On the other hand, using only the false positive loss it tends to predict one huge blob.

\subsubsection{Split Heuristics Analysis.}
In Fig. \ref{fig:speed}  we show that the watershed split achieves better MAE on Trancos and Penguins validation sets. Further, using the watershed split achieves much faster improvement on the validation set with respect to the number of epochs. This suggests that using proper heuristics to identify the negative regions is important, which leaves an open area for future work.



\section{Conclusion}
\label{sec:conclusion}
We propose LC-FCN, a fully-convolutional neural network, to address the problem of object counting using point-level annotations only. We propose a novel loss function that encourages the model to output a single blob for each object instance. Experimental results show
that LC-FCN outperforms current state-of-the-art models on the PASCAL VOC
2007, Trancos, and Penguins datasets which contain objects that are heavily occluded. For future work, we plan to explore different FCN architectures and splitting methods that LC-FCN can use to efficiently split between overlapping
objects that have complicated shapes and appearances. 

\section{Acknowledgements}
We would like to thank the anonymous referees for their useful comments that significantly improved the
paper. Issam Laradji is funded by the UBC Four-Year Doctoral Fellowships (4YF).
\clearpage

\bibliographystyle{splncs}
\bibliography{references}

\end{document}